\documentclass[10pt,twocolumn,letterpaper]{article}

\usepackage{wacv}
\usepackage{times}
\usepackage{epsfig}
\usepackage{graphicx}
\usepackage{amsmath}
\usepackage{amssymb}

\usepackage{mathtools}
\usepackage{varwidth}
\usepackage{newfloat}

\DeclareMathOperator*{\argmax}{arg\,max}
\DeclareMathOperator*{\argmin}{arg\,min}
\usepackage{lipsum}
\usepackage{amsfonts}
\usepackage[outdir=./]{epstopdf}
\usepackage{algorithm}

\usepackage{subcaption}
\usepackage{bbm}
\usepackage{amssymb}
\usepackage{tikz}
\usepackage{relsize}
\usepackage{adjustbox}

\makeatletter
\def\blfootnote{\xdef\@thefnmark{}\@footnotetext}
\makeatother

\usetikzlibrary{shapes.geometric, arrows, fit,decorations.markings,spy}
\tikzstyle{HDim} = [rectangle,minimum width=5cm, minimum height=0.3cm, text centered, draw=black, fill=black]
\tikzstyle{LDim} = [rectangle,minimum width=1cm, minimum height=0.3cm, text centered, draw=black, fill=black]
\tikzstyle{NN} = [rectangle,rounded corners,minimum width=3.25cm, minimum height=3.25cm, text centered, draw=black, fill=blue!30]
\tikzstyle{Cost} = [circle, inner sep = 7.5, draw=black, fill=red!35]
\tikzstyle{textbox} = [rectangle, minimum width=3cm, minimum height=1cm, text centered]
\tikzstyle{arrow} = [thick,->,>=latex, thick,line width=2pt,shorten >= 2pt, shorten <= 2pt]
\tikzstyle{doublearrow} = [thick,<->,>=latex, thick, double,line width=1.75pt]
\tikzstyle{curvebox} = [rectangle, minimum width=3cm, minimum height=1cm, text centered]

\wacvfinalcopy 

\def\wacvPaperID{550} 

\ifwacvfinal\fi
\begin{document}

\title{DIMAL: Deep Isometric Manifold Learning Using Sparse Geodesic Sampling}
\author{Gautam Pai \hspace{1cm} Ronen Talmon \hspace{1cm} Alex Bronstein \hspace{1cm} Ron Kimmel \\
Technion - Israel Institute Of Technology\\
{\tt\small \{paigautam,bron,ron\}@cs.technion.ac.il, ronen@ee.technion.ac.il}}

\maketitle
\ifwacvfinal\thispagestyle{empty}\fi

\begin{abstract}
This paper explores a fully unsupervised deep learning approach for computing distance-preserving maps that generate low-dimensional embeddings for a certain class of manifolds. 
We use the Siamese configuration to train a neural network to solve the problem of least squares multidimensional scaling for generating maps that approximately preserve geodesic distances. 
By training with only a few landmarks, we show a significantly improved local and non-local generalization of the isometric mapping as compared to analogous non-parametric counterparts.
Importantly, the combination of a deep-learning framework with a multidimensional scaling objective enables a numerical analysis of network architectures to aid in understanding their representation power. This provides a geometric perspective to the generalizability of deep learning.
\end{abstract}

\section{Introduction}
The characterization of distance preserving maps is of fundamental interest to the problem of non-linear dimensionality reduction and manifold learning. 
For the purpose of achieving a coherent global representation, it is often desirable to embed the high-dimensional data into a space of low dimensionality while preserving the metric structure of the data manifold. 
The intrinsic nature of the geodesic distance renders such a representation dependent only on the geometry of the manifold and not on how it is embedded to the ambient space. 
In the context of dimensionality reduction, this property makes the resulting embedding meaningful.
\blfootnote {This research is partially supported by the Israel Ministry of Science, grant number  3-14719, the Technion Hiroshi Fujiwara Cyber Security Research Center and the Israel Cyber Directorate. The first author is supported by a fellowship from the Hiroshi Fujiwara Cyber Security Research Center at the Technion.}

The success of deep learning has shown that neural networks can be trained as powerful function approximators of complex attributes governing various visual and auditory phenomena. The hallmark of deep learning has been its ability to automatically learn meaningful representations from raw data without any explicit axiomatic constructions. The availability of large amounts of data and computational power, coupled with parallel streaming architectures and improved optimization techniques, have all led to computational frameworks that efficiently exploit their representational power. However, a study of their behavior under geometric constraints is an interesting question which has been relatively unexplored.   

\begin{figure}[t]
\includegraphics[scale=0.3]{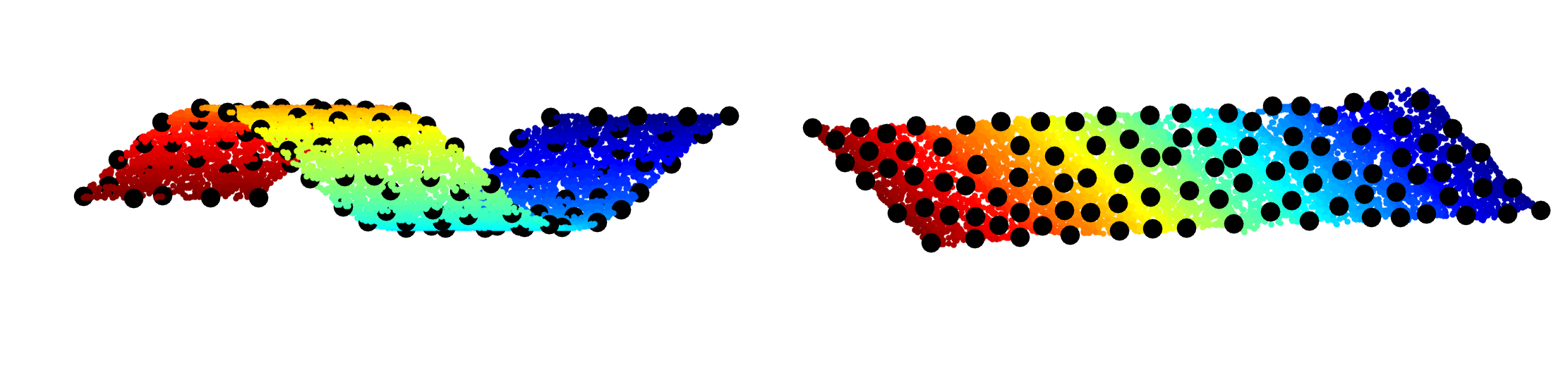}
\vspace{-0.8cm}
{\caption{{\bf \emph{Learning to unfold a ribbon:}} A three dimensional Helical Ribbon (left) and its two dimensional embedding - a planar parallelogram (right) learned using a two-layer MLP. 
The network was trained using estimated pairwise geodesic distances between {\bf only 100} points (marked in black) out of the total 8192 samples.}
\label{HelicalRibbon}}
\end{figure}

In this paper, we use the computational infrastructure of neural networks to model maps that preserve geodesic distances on data manifolds. 
We revisit the classical geometric framework of multidimensional scaling to find a configuration of points that satisfy pairwise distance constraints. 
We show that instead of optimizing over the individual coordinates of the points, we can optimize over the \emph{function that generates these points} by modeling this map as a neural network. 
This choice of modeling the isometric map with a parametric model provides a straightforward out-of-sample extension, which is a simple forward pass of the network. We exploit efficient sampling techniques that progressively select \emph{landmark points} on the manifold by maximizing the spread of their pairwise geodesic distances. We demonstrate that a small amount of these \emph{landmark points} is sufficient to train a network to generate faithful low-dimensional embeddings of manifolds. 
Figure \ref{HelicalRibbon} provides a visualization of our proposed approach. 

Our motivation to integrate deep learning into the classical isometric embedding problem is two-fold. Firstly, this approach allows us to view the concept of \emph{generalization} from a geometric viewpoint. Algorithms that exhibit poor generalizations will yield embeddings that show a clear visual depiction of suboptimal flattening as in Figures \ref{scurve_fps}, \ref{sparse_mds_figure} \ref{LISO_comp}, \ref{fishbowl} and this can be objectively measured using the stress function (Equation \ref{Stress}, which essentially measures the deviation from isometry). We demonstrate qualitatively and quantitatively that deep learning models provide better \emph{local} and \emph{non-local} generalization properties as compared to the axiomatic interpolation and extrapolation formulas of their non-parametric counter-parts. Secondly, the use of geodesic sampling methods enables further analysis, by measuring global stress as a function of the number of samples used for training. Our experiments demonstrate that analogously to numerical algorithms, one can obtain an \emph{order of accuracy} measure for each neural network architecture thereby measuring how efficient a given neural network is in modeling a function. 
\section{Background}
\label{background}
\subsection{Manifold Learning}
Manifold learning is the process of recovering a low-dimensional representation from a possibly non-linear high-dimensional data. The literature on manifold learning is dominated by spectral methods that have a characteristic computational pattern. 
The first step involves the computation of the $k$ nearest neighbors of all $N$ data points. 
Then, an $N \times N$ square matrix is populated using some geometric principle which characterizes the nature of the desired low-dimensional embedding. 
The eigenvalue decomposition of this matrix is then used to obtain the low-dimensional representation of the manifold. Manifold learning techniques such as Laplacian Eigenmaps \cite{belkin2003laplacian}, LLE \cite{roweis2000nonlinear}, HLLE \cite{donoho2003hessian} and Diffusion Maps \cite{coifman2005geometric} are considered to be local methods, since they are designed to minimize some form of local distortion and hence result in embeddings which preserve locality. 
Methods like Isomap \cite{tenenbaum2000global} are considered global because they enforce preserving all geodesic distances in the low-dimensional embedding. 
All spectral techniques are non-parametric in nature and hence do not characterize the map that generates them. 
Therefore, the computational burden of large spectral decompositions becomes a major drawback when the number of data-points is large. Furthermore, out-of-sample extension of the map is a computationally expensive task.   
\subsection{Neural Networks for Manifold Learning}
Examining the ability of neural networks to represent data manifolds has received considerable interest and has been studied from multiple perspectives. 
From the viewpoint of unsupervised parametric manifold learning, one notable approach is based on the metric-learning arrangement of the Siamese configuration \cite{hadsell2006dimensionality, bromley1994signature, chopra2005learning}. 
Similarly, the parametric version of the Stochastic Neighborhood Embedding \cite{maaten2008visualizing} (and its famous variant, $t$-SNE \cite{van2009learning}) is another example of using a neural network to generate a parametric map that is trained to preserve local structure. Variational Autoencoders \cite{kingma2013auto} design generative models of data and fit them to large data-sets. 
However, these techniques demand an extensive training effort requiring large number of training examples in order to generate satisfactory embeddings. In \cite{basri2016efficient}, the authors have argued that neural-networks can efficiently represent manifolds as a monotonic chain of linear segments by providing an architectural construction and analysis; 
\cite{gong2006neural,mishne2017diffusion,chui2016deep} 
use neural networks specifically for solving the out-of-sample extension problem for manifold learning. 
However, their procedure involves training a network to follow a pre-computed non-parametric embedding rather than adopting an entirely unsupervised approach, thereby inheriting some of the deficiencies of the non-parametric methods.
\subsection{Multidimensional Scaling}
Multidimensional scaling (MDS) is a classical algorithm for obtaining the global picture of data using pairwise distances or dissimilarities information. 
The core idea of MDS is to find an embedding configuration $\mathbf{X = \big{[} x_1, x_2, x_3 ...x_N \big{]}}$, such that all pairwise distances measured in the embedded space (typically $\mathbf{||x_i - x_j||}$) are faithful to the given distances $\mathbf{D}_s = \big[ d_{ij}^2\big]$ as much as possible. Therefore, the input to an MDS algorithm is the \emph{distance matrix} and the output is the embedding configuration $\mathbf{X}$ that preserves these pairwise distances. In the context of isometric manifold learning, the MDS framework is enabled by imputing pairwise \emph{geodesic} distances and recovering a low-dimensional embedding output such that pairwise Euclidean distances in this space match the corresponding geodesic distances. Putting simply, this is the action of \emph{flattening} the manifold as shown in Figure \ref{HelicalRibbon}.    

There are two prominent, yet different versions of MDS: \emph{Classical Scaling} and \emph{Least-Squares Scaling}.
Classical Scaling is based on the observation that the double centering of a pairwise squared distance matrix gives an inner-product matrix which can be factored to obtain the desired embedding. 
Therefore, if $\mathbf{H} = \mathbf{I} - \frac{1}{N}\mathbf{11}^T $ is the centering matrix, classical scaling minimizes the strain of the embedding configuration $\mathbf{X}$ and is computed conveniently using the eigen-decomposition of the $N \times N$ matrix $-\frac{1}{2} \mathbf{H} \mathbf{D}_s \mathbf{H}$:  
\begin{eqnarray}
\mathbf{X}^*_{CS} &=& \argmin_{\mathbf{X}} \; || \mathbf{X}\mathbf{X}^T + \frac{1}{2} \mathbf{H} \mathbf{D}_s \mathbf{H} ||^2_F \label{Strain} \\
\mathbf{X}^*_{CS} &=& \mathbf{V \Lambda}^\frac{1}{2}, \; \text{where}\; -\frac{1}{2} \mathbf{H} \mathbf{D}_s \mathbf{H} =  \mathbf{V \Lambda V^T} \label{CS_scaling} 
\end{eqnarray}

At the other end, least squares scaling is based on minimizing the misfits between the pairwise distances of $\mathbf{X = \big{[} x_1, x_2, x_3 ...x_N \big{]}}$ and desired distances $\big[d_{ij}\big]$ measured by the \emph{stress} function
\begin{eqnarray}
\mathbf{X}^*_{LS} &=& \argmin_{\mathbf{X}} \; \sigma(\mathbf{X}) \label{LS_scaling}\\
\sigma(\mathbf{X})   &=& \sum_{i<j} \; w_{ij} \big( ||\mathbf{x}_i - \mathbf{x}_j|| - d_{ij} \big)^2 \label{Stress}
\end{eqnarray}

Minimization of (\ref{Stress}) is typically handled using gradient descent iterations with $w_{ij} = 1$ for all $i,j$. One particular case was introduced by Leeuw \emph{et al.}\cite{de2011multidimensional} under the name SMACOF (Scaling by Majorizing a COmplicated Function). The algorithm is based on the following iterative step:
\begin{equation}
\mathbf{X}_{k+1} = (\mathbf{V}+\frac{1}{N}\mathbf{11}^T)^{\dagger} \; \mathbf{B}(\mathbf{X}_k)\;\mathbf{X}_k
\label{smacof}
\end{equation}
where $N \times N$ matrices $\mathbf{V} = \big[ v_{ij} \big]$ and $\mathbf{B}(\mathbf{X}_k) = \big[ b_{ij} \big]$ are given by:
\begin{eqnarray}
v_{ij} &=&
  \begin{cases}
   -w_{ij} & i\neq j \\
   \sum\limits_{k \neq i} w_{ik} & i=j\\
  \end{cases} \label{Vmatrix}\\ 
b_{ij} &=&
  \begin{cases}
   -w_{ij}\frac{d_{ij}}{\mathbf{||x_i - x_j||}} & i\neq j, x_i \neq x_j \\
   0 & i\neq j, x_i=x_j \\
   -\sum\limits_{k \neq i} b_{ik} & i=j.\\
  \end{cases} \label{Bmatrix}
\end{eqnarray}
The iteration of (\ref{smacof}) guarantees a non-increasing stress, due to the principle of majorization. For more details we refer the interested reader to \cite{de2011multidimensional}.

In practice, the MDS framework is enabled by estimating all pairwise \emph{geodesic} distances with a shortest path algorithm like Dijkstra's \cite{dijkstra1959note}, and choosing an MDS scaling algorithm to generate low-dimensional embeddings that preserve metric properties of the manifold. Schwartz \emph{et al.} \cite{schwartz1989numerical,wolfson1989computing} and the Isomap algorithm \cite{tenenbaum2000global} were the first to suggest populating the inter-geodesic distance matrix $\mathbf{D}_s$ using Dijkstra's algorithm. \cite{grossmann2002computational,zigelman2002texture} first suggested the use of consistent approximate of geodesics for the goal of flattening into 2D and \cite{elad2001bending} employed the Fast Marching method for computing the geodesic distances on 2D surfaces. 

Historically, MDS was first developed in \cite{torgerson1952multidimensional,young1938discussion} who proposed the purely Euclidean model (input distances were euclidian) of the MDS. Later, non-metric versions of MDS were developed in \cite{kruskal1964multidimensional} and \cite{shepard1962analysis} which focused on a generic dissimilarity information rather than plain inter-point distances. Methods like \cite{rosman2010nonlinear} showed an application of the isometric embedding problem to non-simple manifolds with holes and boundaries. In the pattern recognition literature, a version of least squares scaling (\ref{LS_scaling}) is known as Sammon's non-linear mapping (NLM) \cite{sammon1969nonlinear}. Sammon's NLM proposed a straightforward gradient descent on a scaled stress function similar to (\ref{Stress}) for feature extraction from data. Extensions to Sammon's non-linear mapping using artifical neural networks (called SAMANN) was shown in \cite{de1997sammon,mao1995artificial} but with limited exploration and analysis due to the use of plain Euclidean distances instead of their geodesic counterparts. 
\subsection{Sparse Multidimensional Scaling}
One of the major drawbacks of the plain MDS approach is its computational cost and the lack of a principled generalization framework for unseen data. The minimization of objectives (\ref{Strain}) and (\ref{Stress}) demand computing geodesic distances between \emph{all} pairs of points leading to an expensive eigen-decomposition of a dense $N \times N$ matrix as in the case of classical scaling. Similarly, each step of the SMACOF iteration (\ref{smacof}) demands computing pairwise Euclidean distances between \emph{all} pairs of points. 

This motivated various \emph{fast} MDS algorithms that were based on sampling the manifold and computing the pairwise geodesic distances between these sampled pair of points. The input to a \emph{sparse MDS} algorithm is a much smaller $K \times K$ matrix of distances with $K \ll N$ as compared to the entire $N \times N$ matrix required in (\ref{CS_scaling}) and (\ref{smacof}). The embedding of the whole dataset is then extracted using different interpolation philosophies. For example, Aflalo \emph{et al.} \cite{aflalo2013spectral} used spectral geometry of the Laplace Beltrami operator of the manifold to interpolate the distances of the remaining points in a classical scaling framework. Similar in spirit, Boyarski \emph{et al.} \cite{boyarski2017subspace} used spectral interpolation of the SMACOF iterations (\ref{smacof}), minimizing the stress for only a subsampled configuration of points and interpolating the rest using eigenvectors of the Laplace Beltrami operator. Shamai \emph{et al.} \cite{shamai2015classical,shamai2015accelerating} developed a sparse MDS using the Nystrom extension method. A similar approach is also adopted in multigrid-MDS \cite{bronstein2006multigrid} and vector extrapolation \cite{rosman2008fast}.   

Although the use of a secondary interpolation scheme provided a speed up, these frameworks still lack a principled means to generalize to unseen data. The Landmark Isomap method \cite{de2004sparse} along with the generalized formulation in \cite{Bengio+chapter2007} showed an out-of-sample extension using a kernel formula based on computing geodesic distances of the new samples to existing landmarks and triangulating their embedding accordingly. As we show in our experiments in Sections \ref{sec:sparse_mds} - \ref{sec:extensions}, our deep-learning approach outperformes all these methods without the need for any external information such as spectral geometry, Nystrom projection matrices or out-of sample geodesic distances.
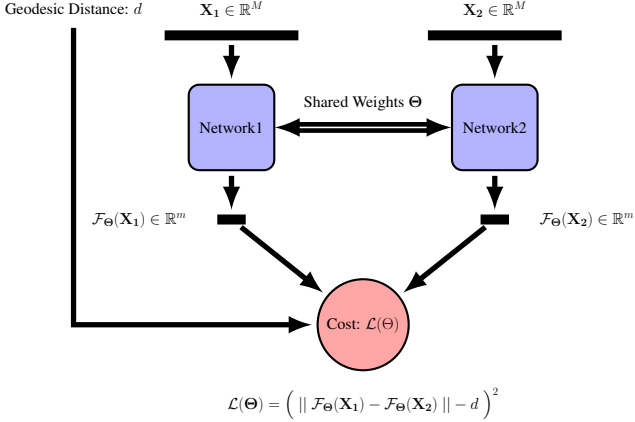
\begin{figure}[h]
\begin{tikzpicture}[thick, every node/.style={scale=0.35,font=\LARGE}]
\node(textLeft)[textbox]{$\mathbf{X_1} \in \mathbb{R}^M$};
\node(textRight)[textbox,right of = textLeft, xshift = 9cm]{$\mathbf{X_2} \in \mathbb{R}^M$};
\node(Label)[textbox, left of = textLeft, xshift = -5cm]{Geodesic Distance: $d$};

\node (curveLeft) [HDim, below of = textLeft] {};
\node (curveRight)[HDim, below of = textRight] {};

\node (NN_Left) [NN, below of  = curveLeft, yshift = -2.5cm] {Network1};
\node (NN_Right) [NN, below of = curveRight, yshift = -2.5cm] {Network2};

\node (SigLeft) [LDim, below of = NN_Left, yshift = -2.5cm] {};
\node (SigRight)[LDim, below of = NN_Right, yshift = -2.5cm] {};

\node(textSigLeft)[textbox, left of = SigLeft,xshift = -2.5cm]{$\mathbf{\mathcal{F}_\Theta (X_1)} \in \mathbb{R}^m$};
\node(textSigRight)[textbox,right of = SigRight,xshift = 2.5cm]{$\mathbf{\mathcal{F}_\Theta (X_2)} \in \mathbb{R}^m$};

\node (Cost) [Cost, below of = SigLeft,yshift  = -3cm, xshift = 5cm]{Cost: $\mathcal{L}(\Theta)$};

\node(MathCost)[textbox, below of = Cost,yshift  = -2cm]{$\mathbf{ \mathcal{L}(\Theta) = \Big(\;||\; \mathcal{F}_\Theta(X_{1}) - \mathcal{F}_\Theta(X_{2})\;||} - d \;\Big)^2$};

\draw [arrow] (NN_Left) -- (SigLeft); 
\draw [arrow] (NN_Right) -- (SigRight);

\draw [arrow] (curveLeft)  -- (NN_Left); 
\draw [arrow] (curveRight) -- (NN_Right);

\draw [arrow] (SigLeft)  -- (Cost); 
\draw [arrow] (SigRight) -- (Cost);

\draw [doublearrow] (NN_Left) -- node [above=0.5cm][anchor=south]{Shared Weights $\mathbf{\Theta}$}(NN_Right);

\draw[arrow] (Label) |- (Cost);
\end{tikzpicture}
\caption{{\bf Siamese Configuration:} Each arm of the siamese network models the map $\mathbf{\mathcal{F}_\Theta}: \mathbb{R}^M \to \mathbb{R}^m,\;\;m \ll M$. For each training pair $(\mathbf{X}_1, \mathbf{X}_2) \in \mathbb{R}^M$, the loss minimizes the squared difference between the Euclidean distances of the low-dimensional embedding and the geodesic distance $d$.}
\label{siameseconfig}
\vspace{-0.3cm}
\end{figure}
\section{DIMAL: Deep Isometric MAnifold Learning}
\label{sec:dimal}
\subsection{Training Configuration}
We incorporate the ideas of least squares scaling into the computational infrastructure of the Siamese configuration as shown in Figure \ref{siameseconfig}. A Siamese configuration comprises of two identical networks that process two different inputs. The outputs are then combined in a loss that is typically a function of the distance between the output pairs. This configuration has been extensively used for the purposes of metric learning, descriptor learning, 3D shape correspondence etc. \cite{bromley1994signature,chopra2005learning,masci2015geodesic}.

For every $p^{th}$ pair of datapoints, we estimate the geodesic distance using a shortest path algorithm and train the network by minimizing the network-parameterized stress function (see Figure \ref{siameseconfig}): 
\begin{equation}
\mathcal{L}(\Theta) = \sum_{p} \; (\;||\; \mathcal{F}_\Theta(\mathbf{X}_{1}^{(p)}) - \mathcal{F}_\Theta(\mathbf{X}_{2}^{(p)})\;|| - d^{(p)}\;)^2
\label{loss_MDSNet}
\end{equation}
The expectation here is that with a sufficient number of example pairs, the network learns to model a map $\mathbf{\mathcal{F}_\Theta}: \mathbb{R}^M \to \mathbb{R}^m,\;\;m \ll M$, from high to low dimension, that preserves isometry. As explained in Section \ref{sec:orrofacc}, we can study exactly how many examples are needed to learn this map to sufficient accuracy, by employing a sampling strategy that sufficiently covers the entire manifold  uniformly.  
\subsection{Geodesic Farthest Point Sampling}
\begin{figure}[h]
\includegraphics[scale=0.28]{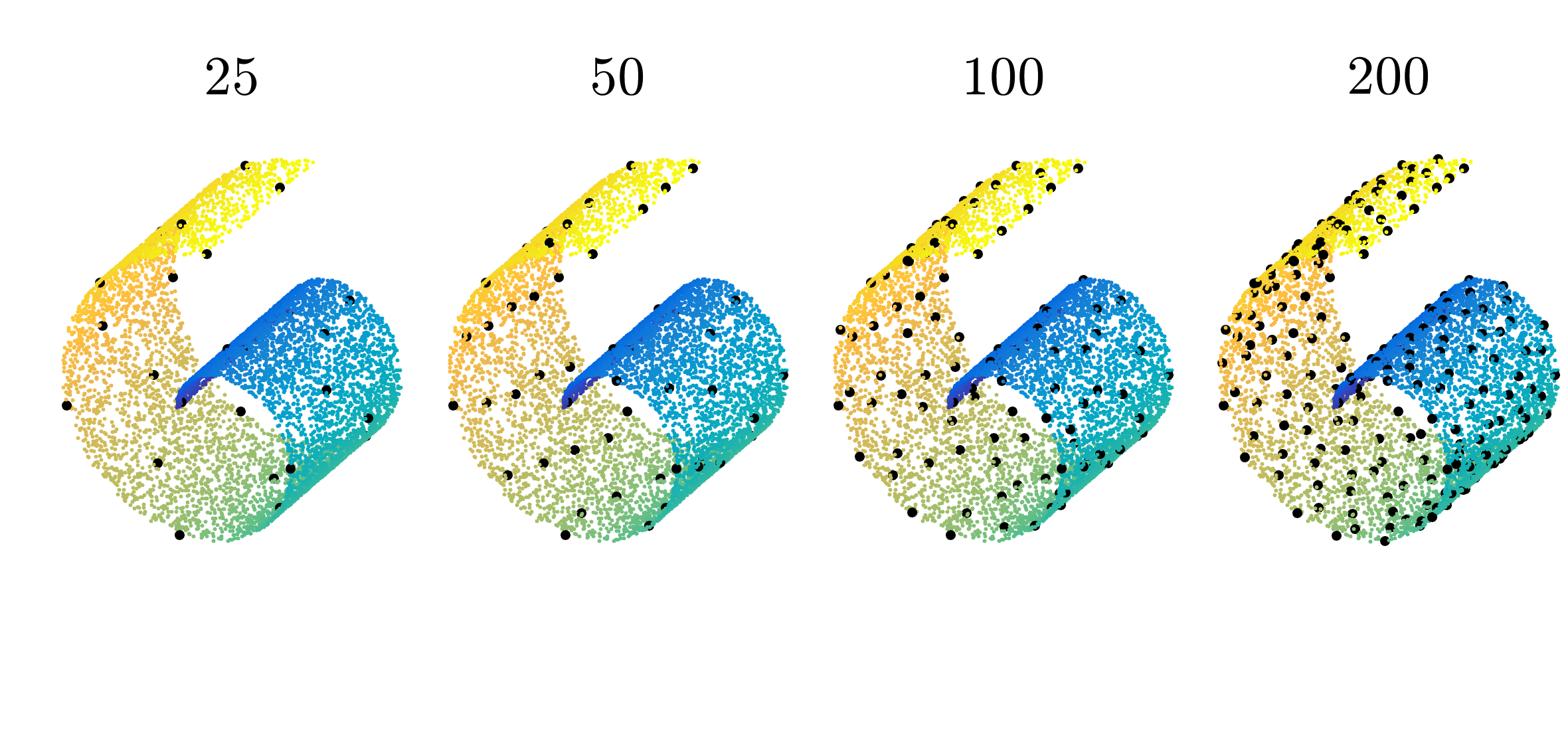}
\vspace{-1cm}
{\caption{{\bf Visualizing Farthest Point Sampling:} The Swiss-roll manifold sampled using the farthest point sampling algorithm (landmarks in black).} 
\label{swissroll}}
\vspace{-0.3cm}
\end{figure}

The farthest point sampling strategy \cite{hochbaum1985best,bronstein2008numerical} (also referred to as the MinMax strategy in \cite{de2004sparse}) is a method for picking landmarks amongst the points of a discretely sampled manifold such that under certain conditions, these samples cover the manifold as uniformly as possible. 
Starting from a random selection, the landmarks are chosen one at a time such that each new selection from the unused samples has the largest geodesic distance to the set of the selected sample points. 
Figure \ref{swissroll} provides a visualization of this sampling mechanism. 
We train the network by minimizing the loss (\ref{loss_MDSNet}) by computing the pairwise geodesic distances between $K$ landmarks. 
Therefore, the pre-training computational effort is limited to $\mathcal{O}(K^2)$ distances. 
\begin{algorithm}[H]
\centering
\begin{tabular}{ll}
{\bf Input:}&metric space $(X,d_X)$, \\ & initial point $x_1\in X$,\\ 
& number of landmarks: K \\
{\bf Output:}&sampling $X' = \{ x_1, x_2, x_3, ...x_K \}$ \\
{\bf Initialization:} &$X' = \{ x_1\},\;d(x) = d_X(x,x_1)$
\end{tabular}\\
\begin{enumerate}
\item [] {\bf While $|X'|<K$ do:} 
\item Find the farthest point from $X'$, $x' = \underset{x\in X}{\argmax}\;d(x)$
\item Update set of selected samples: $X'\longleftarrow X'\cup \{x'\}$
\item Update distance function $d(x)$: $d(x) \longleftarrow \underset{x}{\textrm{min}} \{ d(x),d_X(x,x')$ \}
\item[]  {\bf End}
\end{enumerate}
\caption{Farthest Point Sampling}
\label{algo_fps}
\end{algorithm}
Given a dataset with $N$ samples, we can summarize the proposed DIMAL algorithm as follows (see Algorithm \ref{algo_dimal}). We first build a graph from the data using an approximate nearest neighbor algorithm, and obtain the set of landmarks and corresponding pairwise geodesic distances using farthest point sampling (Algorithm \ref{algo_fps}). Then, we construct a dataset of all pairs of landmarks and their corresponding geodesic distances. Using the Siamese configuration depicted in Figure \ref{siameseconfig} we train the network parameters minimizing the MDS loss (\ref{loss_MDSNet}). The low-dimensional embedding for any high-dimensional datapoint is obtained with its forward pass through the trained network.  
\section{Experiments}
The core idea behind our experiments is to train on the sub-sampled landmarks and observe the effect on the entire set. Put in another way, we train to preserve only $\frac{1}{2}K(K-1)$
 distances between the landmarks and evaluate the stress (\ref{Stress}) over \emph{all} $\frac{1}{2}N(N-1)$ datapoints. We perform three separate experiments using this evaluation methodology. First, in Section \ref{sec:orrofacc} we vary the number of landmarks $K$, and observe the overall stress as a function of the network architecture. Second, in Section \ref{sec:sparse_mds} we compare DIMAL to existing sparse MDS frameworks by imputing all the algorithms with the \emph{same landmarks and same geodesic distances} and evaluating the MDS outputs using overall stress. Third, in Section \ref{sec:l_isomap} we analyze the out-of-sample performance of our framework and compare it to the only axiomatic extension of MDS: Landmark Isomap \cite{silva2003global}. 
Finally, we show an extension of our framework to non-isometric manifolds in Section \ref{sec:extensions}. 
\begin{algorithm}[H]
\begin{tabular}{ll}
{\bf Input:}& High-dimensional data-points of manifold:\\ 
&$\{\mathbf{X}_i \in \mathbb{R}^M\}$, $i=1,2,...N$,\\ 
& number of landmarks: K \\
{\bf Output:}& Low-dimensional embeddings of datapoints:\\
&$\{\mathbf{x}_i \in \mathbb{R}^m, m\ll M \}$, $i=1,2,...N$, \\
\end{tabular}\\
\begin{enumerate}
\setlength\itemsep{0.7em}
\item Compute the nearest-neighbor graph from the manifold data and obtain a set of $K$ landmark	points using Algorithm \ref{algo_fps}
\item Obtain pairwise geodesic distances $\mathbf{D}_s$ between landmarks using Dijkstra's or any other numerical algorithm 
\item Form a dataset of landmark pairs with corresponding geodesic distances $\bigg\{ \mathbf{X}_1^{(p)}, \mathbf{X}_2^{(p)}, d^{(p)} = d(\mathbf{X}_1^{(p)},\mathbf{X}_2^{(p)}) \bigg\}$
\item {\bf Training:} Train network $\mathcal{F}_\Theta$ with Siamese configuration of Figure \ref{siameseconfig} minimizing loss (\ref{loss_MDSNet}) for the dataset obtained in Step 3
\item {\bf Inference:} Obtain low-dimensional embedding with forward pass of the network: $\mathbf{x}_i = \mathcal{F}_\Theta(\mathbf{X}_i), \; \forall i\in \{1,2,...N\}$  
\end{enumerate}
\caption{DIMAL: Deep Isometric MAnifold Learning}
\label{algo_dimal}
\end{algorithm}
\begin{figure*}
\begin{center}
\hspace{-1.3cm}
\begin{subfigure}{0.37\textwidth}
\includegraphics[scale=0.35]{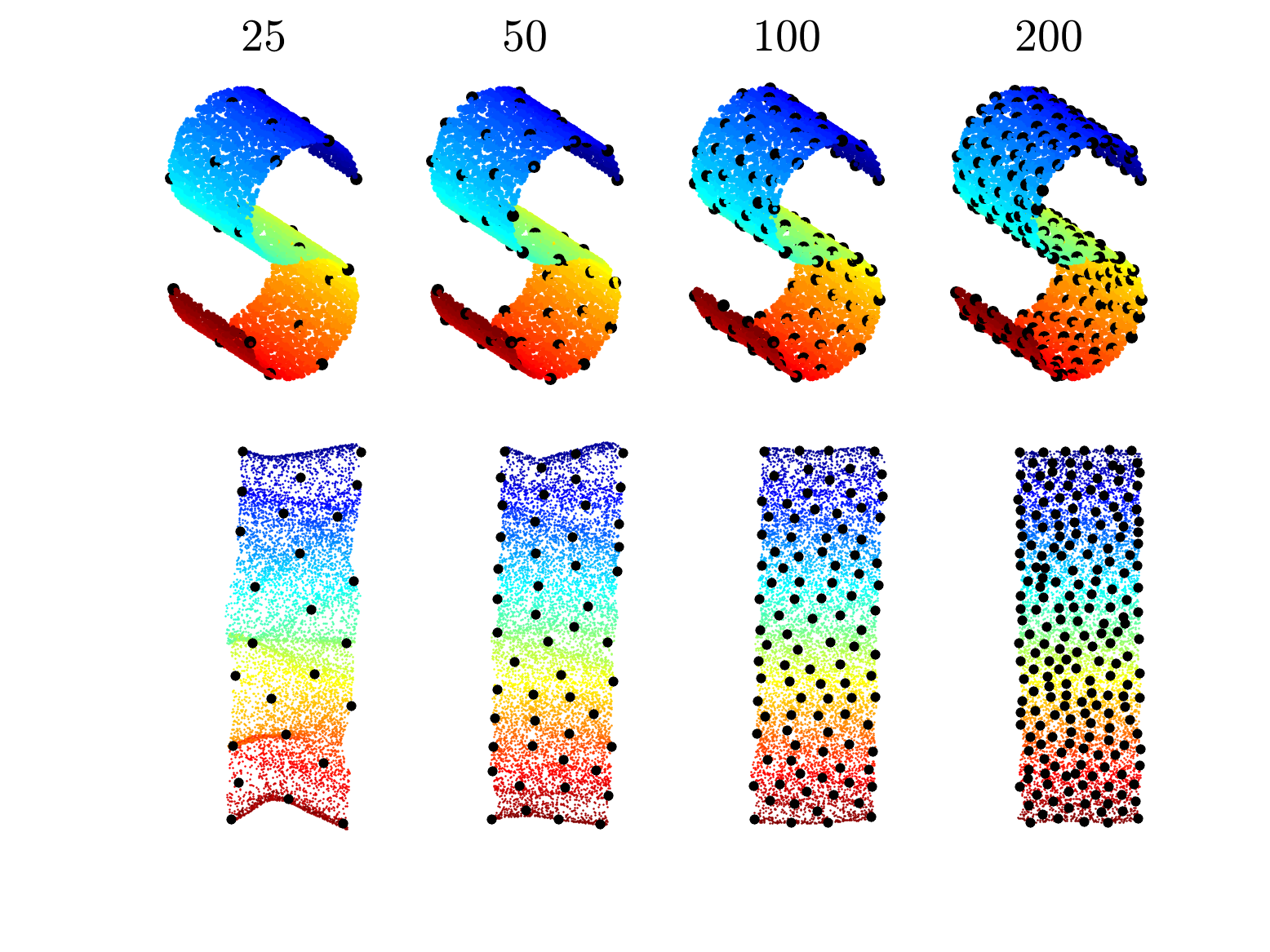}
\vspace{-1.1cm}
\caption{}
\label{scurve_fps}
\end{subfigure}
\begin{subfigure}{0.31\textwidth}
\includegraphics[scale=0.22]{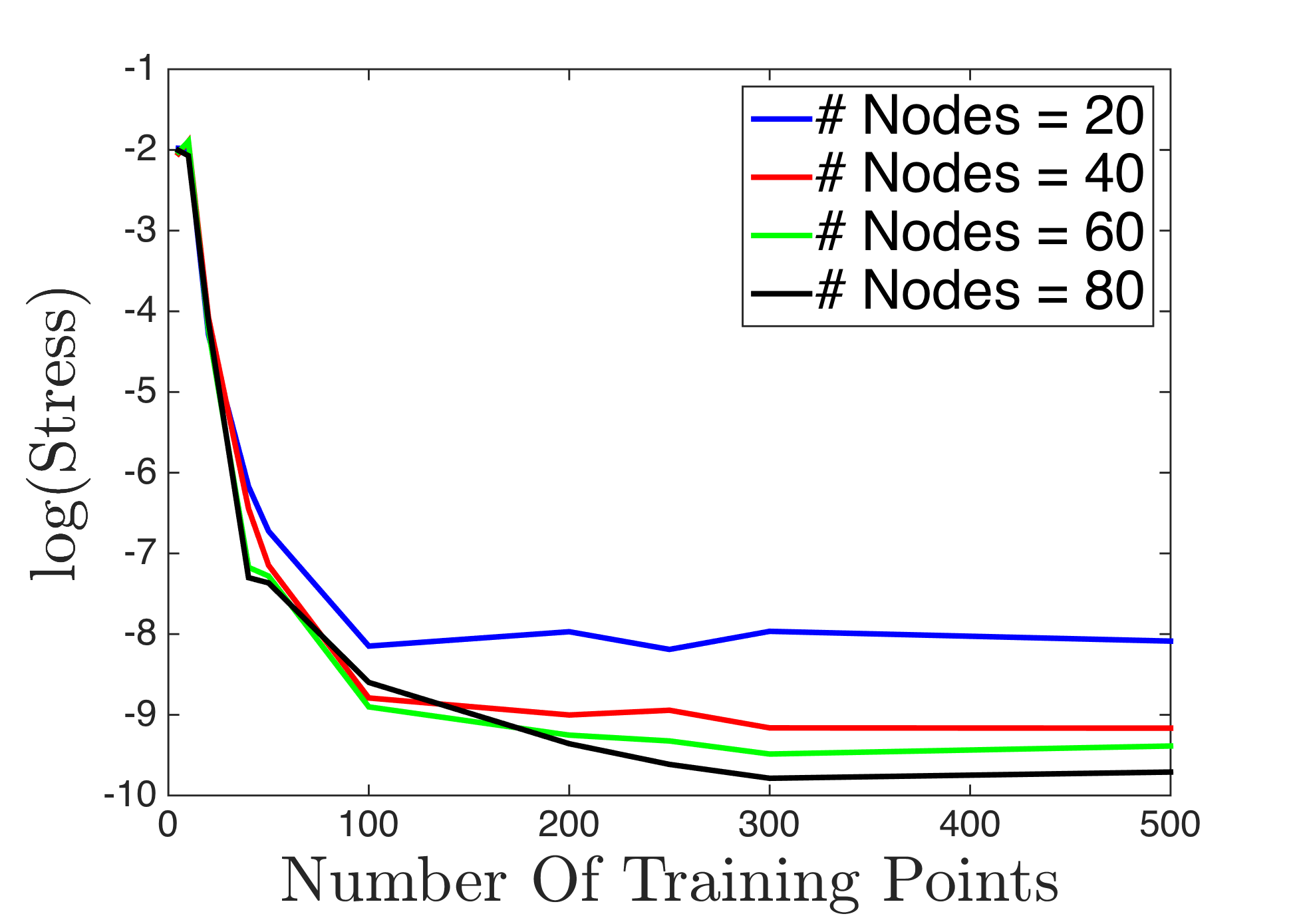}
\caption{}
\label{num_exp_1}
\end{subfigure}
\begin{subfigure}{0.31\textwidth}
\includegraphics[scale=0.22]{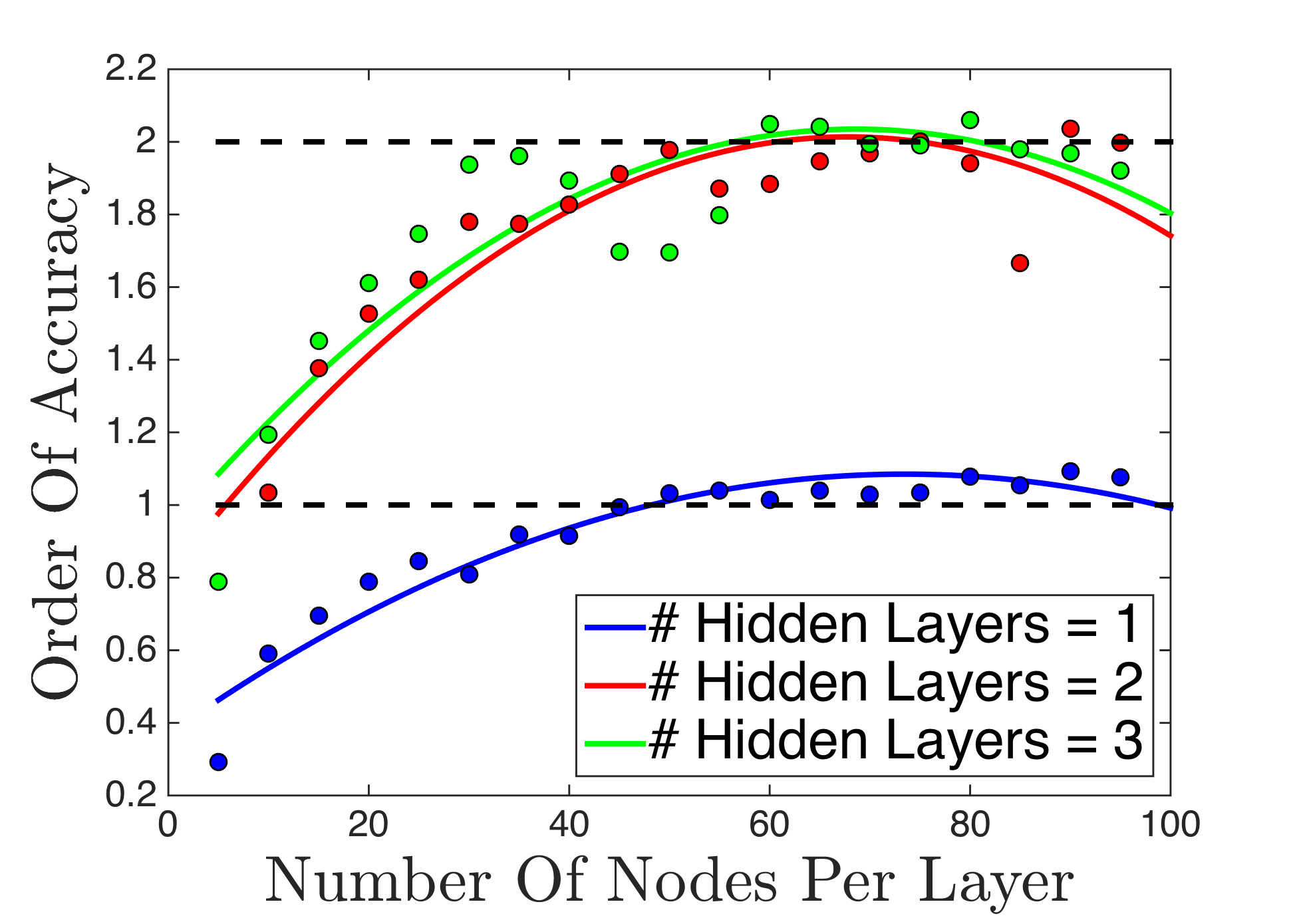}
\caption{}
\label{num_exp_2}
\end{subfigure}
\end{center}
\vspace{-0.5cm}
\caption{{\bf Exploring the variations in architecture and the number of landmarks:}
(a) (top row) Three dimensional S-Curve manifold with varying number of landmark points obtained using Algorithm \ref{algo_fps}. (bottom row) The corresponding two dimensional DIMAL output (Algorithm \ref{algo_dimal}) generated by a 2-Layer MLP with 70 hidden nodes per layer.
(b) The logarithm of stress of all 8172 points as a function of number of landmarks. 
(c) Order of accuracy estimates of varying architectures.}
\label{num_exp}
\end{figure*}
\subsection{Numerical Experiments on 3D Point Clouds}
\label{sec:orrofacc}
Our first set of experiments is based on synthetic point-cloud manifolds, like S-curve (Figure \ref{scurve_fps}) and the Helical ribbon (Figure \ref{HelicalRibbon}). 
We use a multilayer perceptron (MLP) with PReLU
\begin{equation}
\text{PReLU}(x)=\max(0,x)+a \; \min(0,x),
\end{equation}
as the non-linear activation function, where $a$ is a learnable parameter. 
The networks are trained for 1000 iterations using the ADAM optimizer with constants $(\beta_1, \beta_2) = (0.95,0.99)$ and a learning rate of 0.01. 
We run each optimization 5 times with random initialization to ensure convergence. 
All experiments were implemented in Python using the PyTorch framework \cite{paszke2017pytorch}. 
We used the scikit-learn machine learning library for the nearest-neighbor and scipy-sparse for Dijkstra's shortest path algorithms.      

Figure \ref{HelicalRibbon} and \ref{scurve_fps} show the results of our method on the Helical ribbon and S-curve respectively with varying number of training samples (in black) out of a total of 8172 data points. 
The number of landmarks dictates the approximation quality of the low-dimensional embedding generated by the network. 
Training with too few samples results in inadequate generalization which can be inferred from the corrugations of the unfolded manifold embedding in the first two parts of Figure \ref{scurve_fps}. Increasing the number of landmarks improves the quality of the embedding, as expected. 
We compute the stress function (\ref{Stress}) of the entire point configuration to measure the quality of the MDS fit. 
Figure \ref{num_exp_1} shows the decay in the stress as a function of the number of training points (landmarks) in a two layer MLP. 

The natural questions to ask are \emph{how many landmarks? how many layers? and how many hidden nodes per layer?} 
We observe that these questions relate to an analogous setup in numerical methods for differential equations. 
For a given numerical technique, the accuracy of the solution depends on the resolution of the spatial grid over which the solution is estimated. 
Therefore, numerical methods are ranked by an assessment of the \emph{order of accuracy} their solutions observe \cite{levequefinite}.  
This can be obtained by assuming that the relationship between the approximation error $E$ and the resolution of the grid $h$ is given by
\begin{equation}
E = C\;h^P
\label{errorfn}
\end{equation}
where $P$ is the order of accuracy of the technique and $C$ is some constant. Thus, for every spatial resolution $h$, the error of the numerical algorithm $E(h)$ is evaluated and $\log E(h)$ is plotted as a function of $\log h$ for the specific algorithm. $P$ is obtained by computing the slope of the line since,
\begin{equation}
\log(E) = \log(C) + P\;\log(h).
\label{errorfnlog}
\end{equation}

We extend the same principle in order to evaluate network architectures (in place of numerical algorithms) for estimating the quality of isometric maps. 
We use the overall stress (\ref{Stress}) as the error function $E$ in (\ref{errorfn}). 
We assume that due to the 2-optimal property \cite{hochbaum1985best} of the farthest point strategy for a two dimensional manifold, the sampling is approximately uniform and hence $h \propto \frac{1}{\sqrt{K}}$ where $K$ is the number of landmarks.
By varying the number of layers and the number of neurons per layer, we associate an order of accuracy to each architecture using (\ref{errorfnlog}), by training with a varying number of landmarks $K$ and evaluating the overall stress. 

Figure \ref{num_exp_2} shows the results of the described experiment. 
It shows that a single layer MLP has the capacity of modeling functions to the first order of accuracy. Adding a layer increases the representation power by moving to a second order result. Adding more layers does not provide any substantial gain arguably due to a larger likelihood of over-fitting as seen in the considerably noisier estimates (in green). Therefore, a two layer MLP with 70 hidden neurons per layer can be construed as a good architecture for approximating the isometric map of the S-Curve of Figure \ref{scurve_fps} with 200 landmarks. 
\subsection{Image Articulation Manifolds}
\label{sec:image_art}
Each point on a non-linear articulation manifold is a binary image that is generated by the articulation of a few parameters. In \cite{donoho2005image} it is shown that certain types of such manifolds are isometric to Euclidean space, that is, the geodesic distance between any two sample points is equal to the Euclidean distance between their articulation parameters. Therefore, one can consider such manifolds to be the multidimensional equivalents of the three dimensional Swiss-roll or S-curve. 
\begin{figure}[h]
\begin{center}
\includegraphics[width=0.55\linewidth]{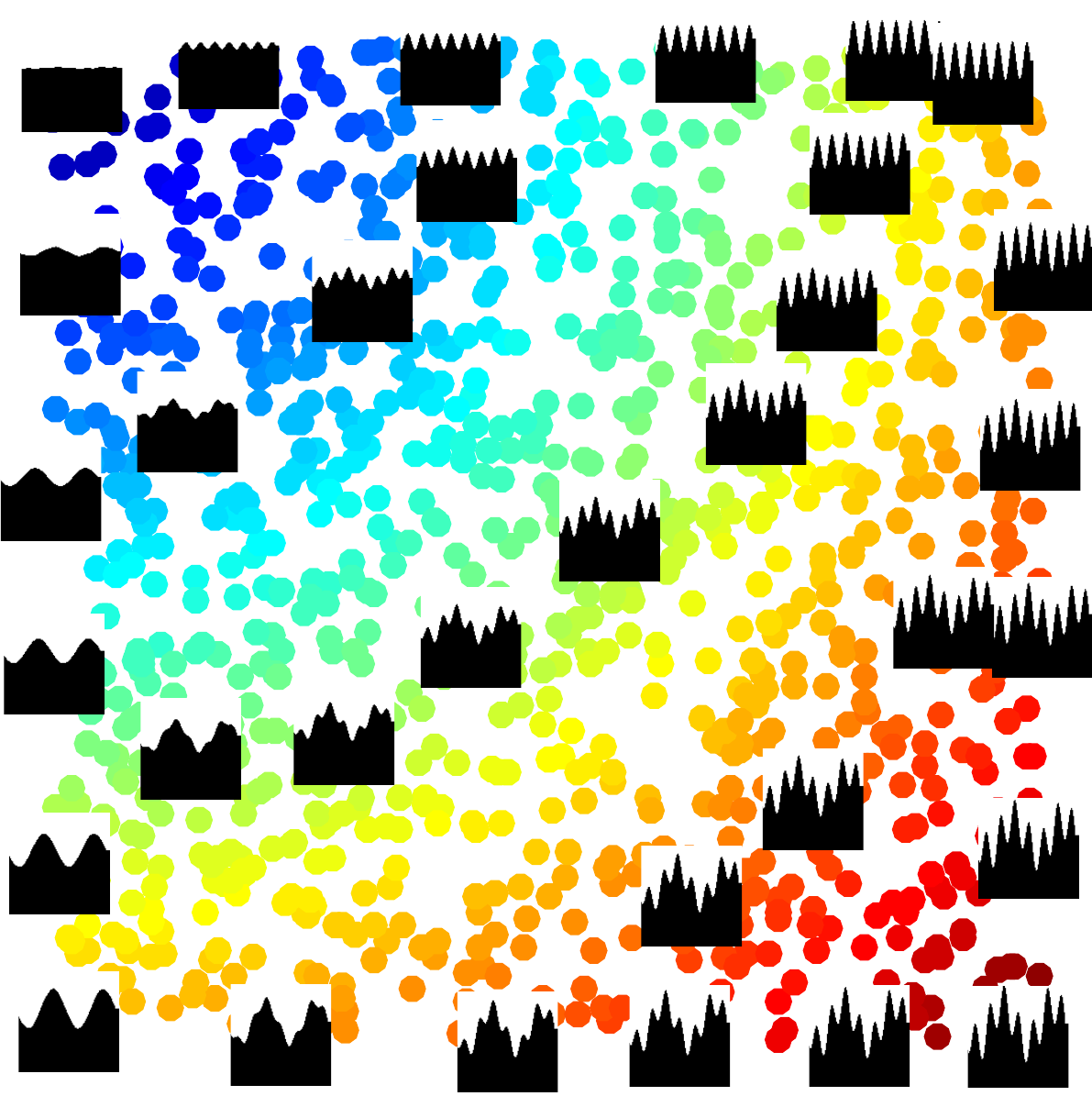}
\end{center}
\vspace{-0.5cm}
\caption{{\bf Visualizing a horizon articulation manifold:} samples generated from the image articulation manifold as per equations (\ref{eq_imagemanifold_1}) and (\ref{eq_imagemanifold_2}) with $\omega_1 = 2$, $\omega_2 = 7$. The color is proportional to the magnitude $\sqrt{\alpha_1^2+\alpha_2^2}$}
\label{image_sinusoids}
\end{figure}
\begin{figure}[h]
\begin{center}
\includegraphics[width=0.8\linewidth]{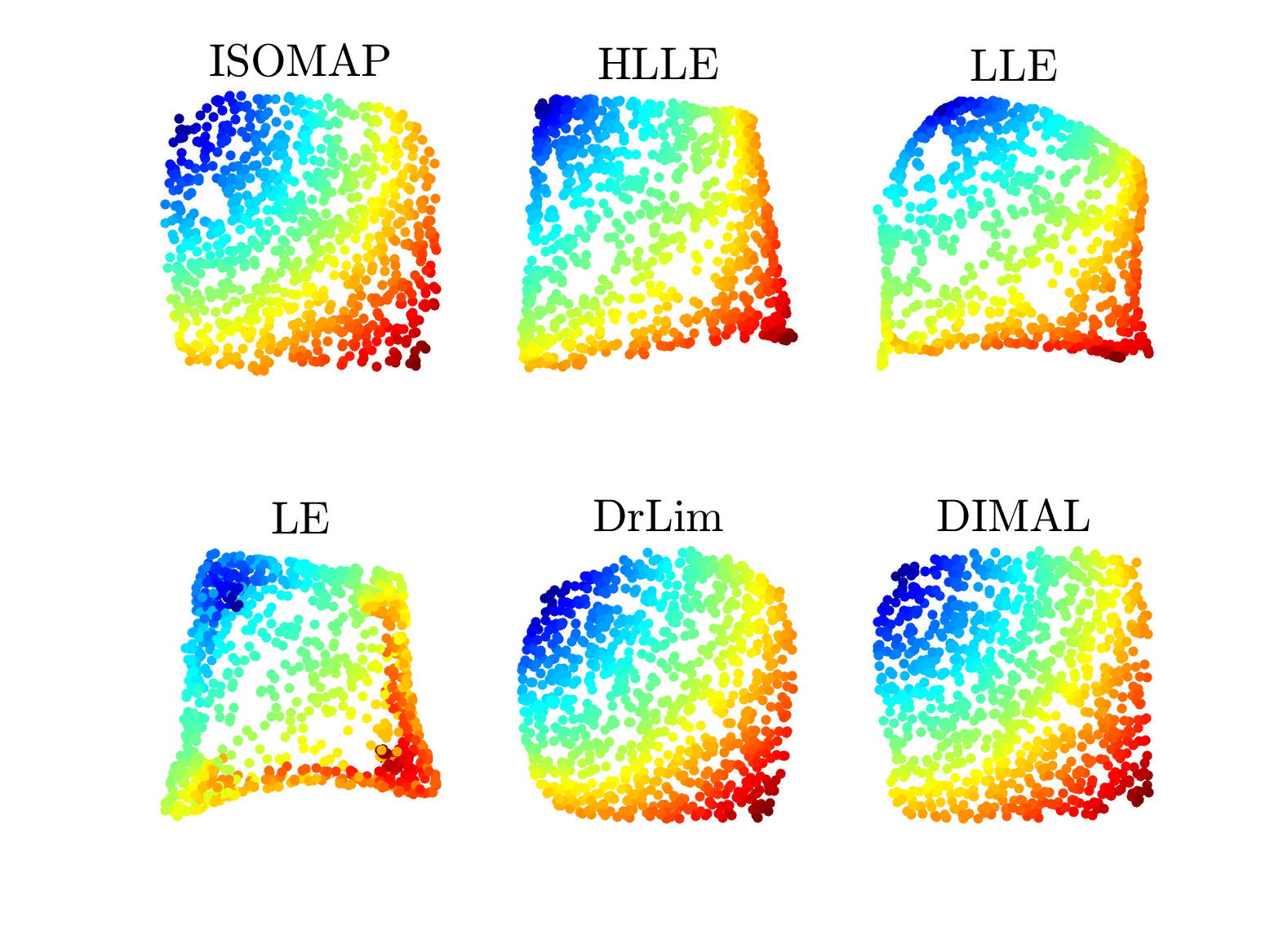}
\end{center}
\vspace{-1cm}
\caption{Comparing metric preservation properties for different manifold learning algorithms on the image articulation manifold dataset. Isomap \cite{tenenbaum2000global}, HLLE \cite{donoho2003hessian}, LLE \cite{roweis2000nonlinear}, LE \cite{belkin2003laplacian}, DrLim \cite{hadsell2006dimensionality}. The proposed method shows maximum fidelity to the ground truth shown in Figure \ref{image_sinusoids}.}
\label{image_sinusoids_comparison}
\end{figure}
We construct a horizon articulation manifold where each image contains two distinct regions separated by a horizon which is modulated by a linear combination of two fixed sinusoidal basis elements as depicted in Figure \ref{image_sinusoids}.  
\begin{eqnarray}
I_{\alpha_1,\alpha_2}(u,v)  &=& \mathbbm{1}_{\{v\leq\psi_{\alpha_1,\alpha_2}(u)\}} \label{eq_imagemanifold_1} \\
\psi_{\alpha_1,\alpha_2}(u) &=& \alpha_1 \sin(\omega_1 u) + \alpha_2 \sin(\omega_2 u)
\nonumber
\end{eqnarray}
Thus, each sample has an intrinsic dimensionality of two - the articulation parameters $(\alpha_1,\alpha_2)$ which govern how the sinusoids representing the horizon are mixed. We sample the articulation parameters from a 2D uniform distribution
\begin{equation}
(\alpha_1,\alpha_2) \sim U([0,1] \times [0,1]).
\label{eq_imagemanifold_2}
\end{equation}

In the context of the main narrative of this paper, which is metric preserving properties of manifolds, we find that such a dataset provides an appropriate test-bed for evaluating metric preserving algorithms. 
Since we are assured of isometry to Euclidean plane, we can objectively measure the performance of an MDS flattening algorithm with the stress function (\ref{Stress}). 
Figure \ref{image_sinusoids_comparison} shows the comparison between DIMAL and other prominent manifold learning algorithms. Except for DIMAL and Isomap, all other methods exhibit some form of distortion indicating a suboptimal metric preservation. 
\begin{figure}[t]
\begin{center}
\includegraphics[scale=0.25]{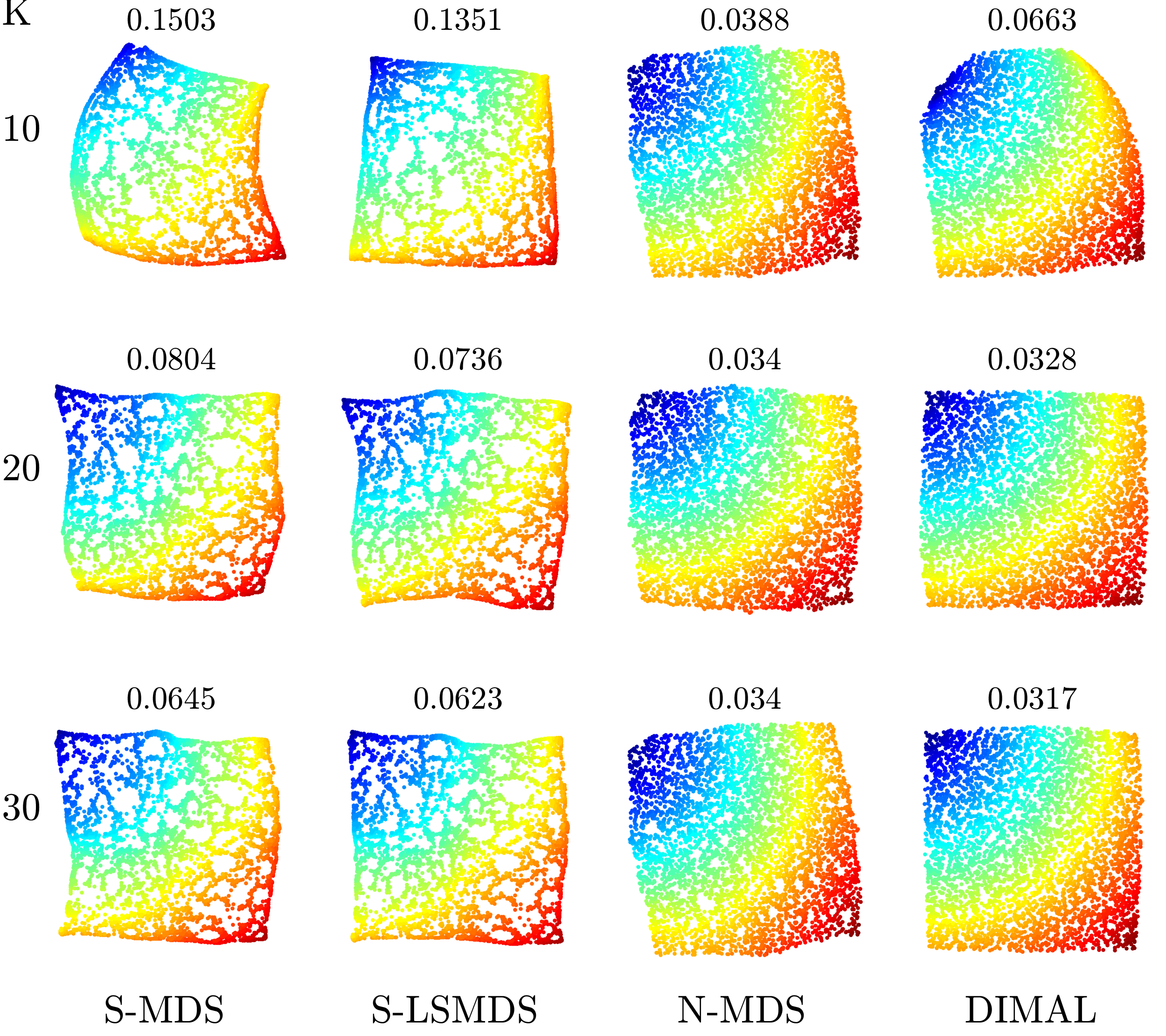}
\end{center}
   \caption{{\bf Comparison with sparse MDS algorithms:} Visual evaluation of the interpolation behavior of different sparse MDS frameworks. The rows are in increasing order of landmarks. The titles denote the corresponding stress of that embedding.}
\label{sparse_mds_figure}
\end{figure}
\begin{figure}[t]
\begin{center}
\includegraphics[scale=0.3]{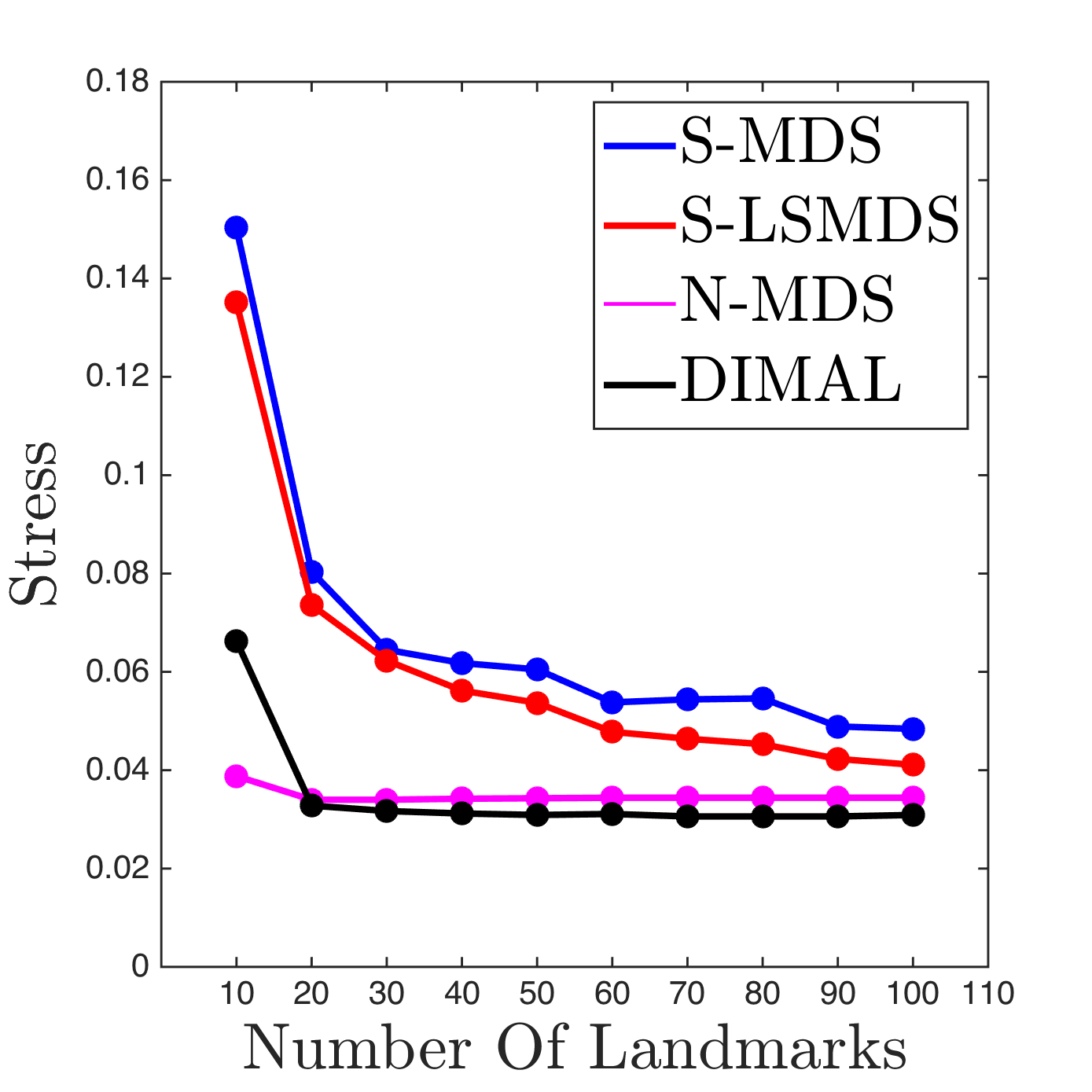}
\end{center}
   \caption{Stress plots as a function of the number of landmarks $K$}
\label{sparse_mds_plot}
\end{figure}
\subsection{Evaluation with Sparse MDS Algorithms:}
\label{sec:sparse_mds}
We compare DIMAL to existing state-of-the-art sparse MDS algorithms and present the results in Figures \ref{sparse_mds_figure} and \ref{sparse_mds_plot}. We generate 5000 examples of the articulation manifold of Figure \ref{image_sinusoids} with $\omega_1 = 2$ and $\omega_2 = 4$. DIMAL was trained with a CNN comprising two convolution layers, each with kernel sizes 12 and 9, number of kernels 15 and 2, respectively, along with a stride of 3 and, followed by a fully-connected layer mapping the image to a two-dimensional domain. We train the network for 500 iterations using the ADAM optimizer \cite{kingma2014adam} with a learning rate of 0.01 and parameters $(\beta_1, \beta_2) = (0.95,0.99)$. 
Each algorithm has been imputed with the \emph{same landmarks} and the \emph{same corresponding pairwise geodesic distances}. For every $K$ landmarks used, we use $\frac{1}{2}K$ eigenvectors of the Laplace-Beltrami operator for the spectral MDS algorithms as suggested in \cite{boyarski2017subspace}. We observe that DIMAL for almost all values of $K$ performs visually and quantitatively better that other MDS algorithms without using any external information. 
\begin{figure}[h]
\begin{center}
\begin{tikzpicture} [thick, every node/.style={scale=0.3,font=\LARGE}, spy using outlines={circle,black,magnification=4,size=5cm, connect spies}]
\node(Emb_MDSNet)[curvebox,anchor=south west,inner sep=0] at(0,0) {\includegraphics[scale = 0.65] {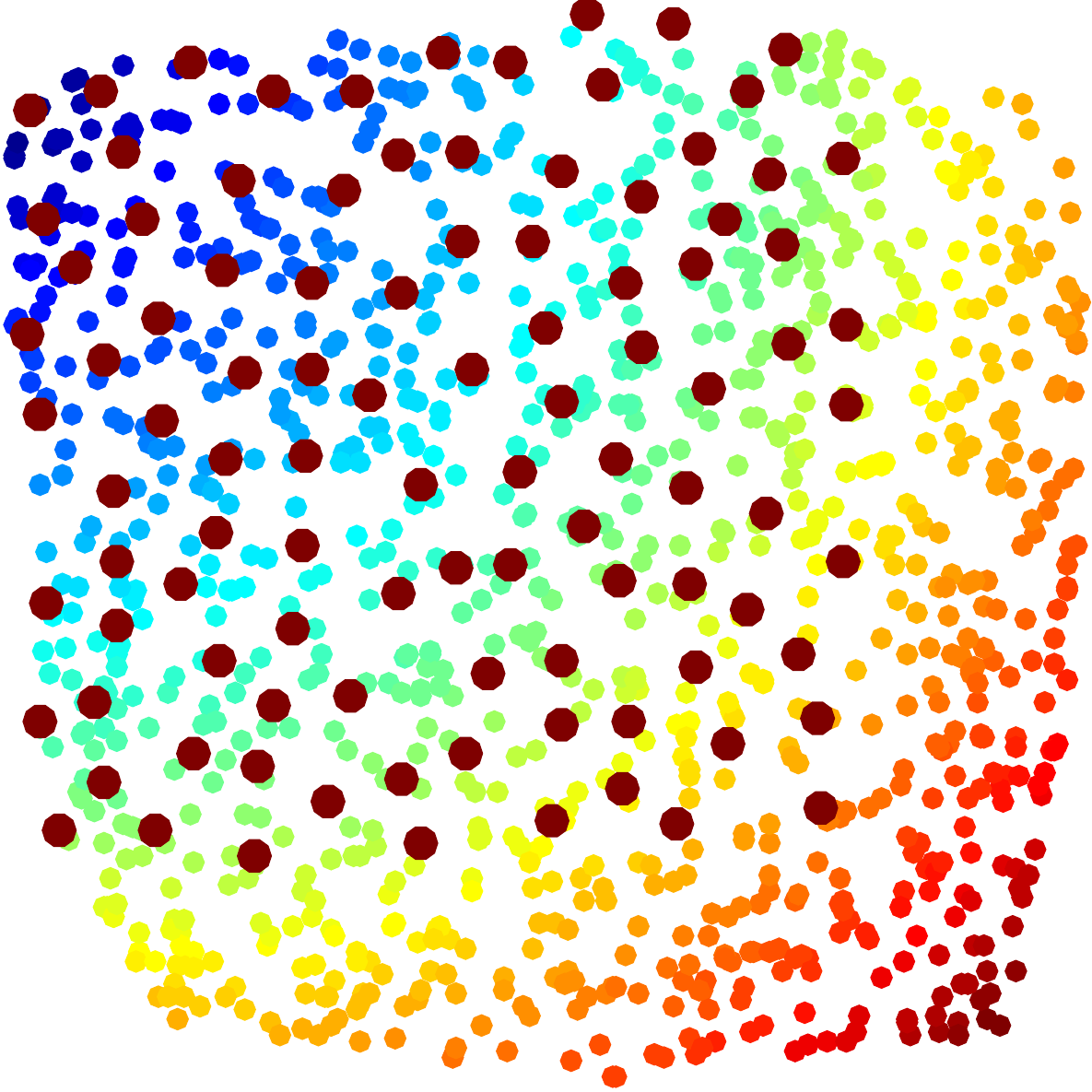}};
\begin{scope}[x={(Emb_MDSNet.south east)},y={(Emb_MDSNet.north west)}]
\coordinate (posspy) at (1.35,.6);
\coordinate (center) at (0.78,.23);
\spy on (center) in node [] at (posspy);
\end{scope}
\node(Caption)[textbox,below of = Emb_MDSNet,xshift = 3cm, yshift = -5cm]{{\Huge DIMAL}};
\end{tikzpicture}
\begin{tikzpicture} [thick, every node/.style={scale=0.3,font=\LARGE}, spy using outlines={circle,black,magnification=4,size=5cm, connect spies}]
\node(Emb_LISO)[curvebox,anchor=south west,inner sep=0] at(0,0) {\includegraphics[scale = 0.65] {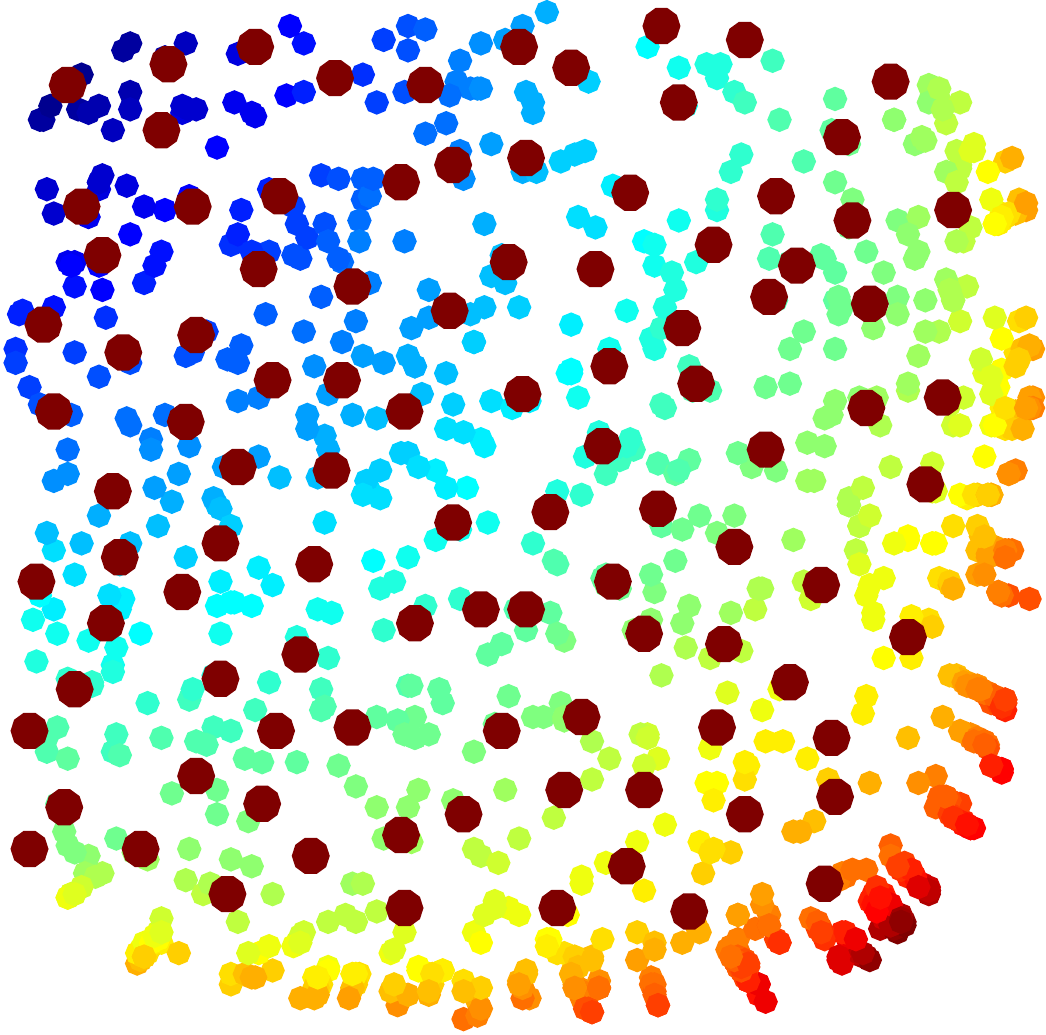}};
\begin{scope}[x={(Emb_LISO.south east)},y={(Emb_LISO.north west)}]
\coordinate (posspy) at (1.4,.6);
\coordinate (center) at (0.77,.2);
\spy on (center) in node [] at (posspy);
\end{scope}
\node(Caption)[textbox,below of = Emb_MDSNet,xshift = 3cm, yshift = -5cm]{{\Huge Landmark Isomap}};
\end{tikzpicture}
\end{center}
\caption{{\bf Evaluation of out-of-sample extensions:} Visualizing the non-local generalization properties of our method (top) and Landmark Isomap (bottom). Both algorithms were trained on the same Landmarks (in red) sampled from only a part of the manifold.}
\label{LISO_comp}
\end{figure}
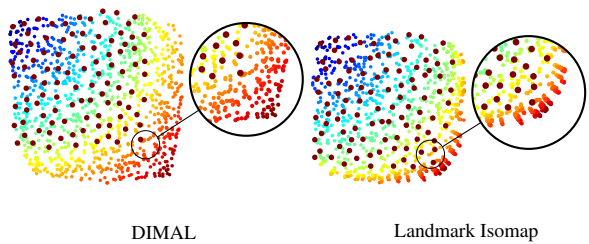
\subsection{Comparison with Landmark Isomap}
\label{sec:l_isomap}
We compare DIMAL to its direct non-parametric competitor: Landmark-Isomap \cite{de2004sparse}. 
The main idea of Landmark-Isomap is to perform classical scaling on the inter-geodesic distance matrix of only the landmarks and then to estimate the embeddings of the remaining points using an interpolating formula (also mentioned in \cite{bengio2004out}). The formula uses the estimated geodesic distances of each new point to the selected landmarks in order to estimate its low dimensional embedding. 

We generate a training horizon articulation dataset containing 5000 samples generated with parameters sampled from $(\alpha_1,\alpha_2) \sim U([0,0.75] \times [0,0.75])$ and evaluate the outputs on test dataset also of 5000 samples with parameters sampled from $(\alpha_1,\alpha_2) \sim U([0,1] \times [0,1])$, thereby isolating a part of the manifold during training.  As in Section \ref{sec:sparse_mds}, both the methods are imputed with the same set of landmarks for evaluation. 

As depicted in Figure \ref{LISO_comp}, the output of Landmark-Isomap shows a clustered result due to the lack of non-local data in the geodesic distance calculations for the interpolation. In contrast, our neural network clearly exhibits a better generalization property, even for parts of the manifold that were isolated during training.
\subsection{Extensions}
\label{sec:extensions}
\subsubsection{Conformal Isometric Mapping}
\begin{figure}[t]
\vspace{-0.5cm}
\begin{center}
\begin{subfigure}[h]{0.5\textwidth}
\centering
\includegraphics[scale=0.27]{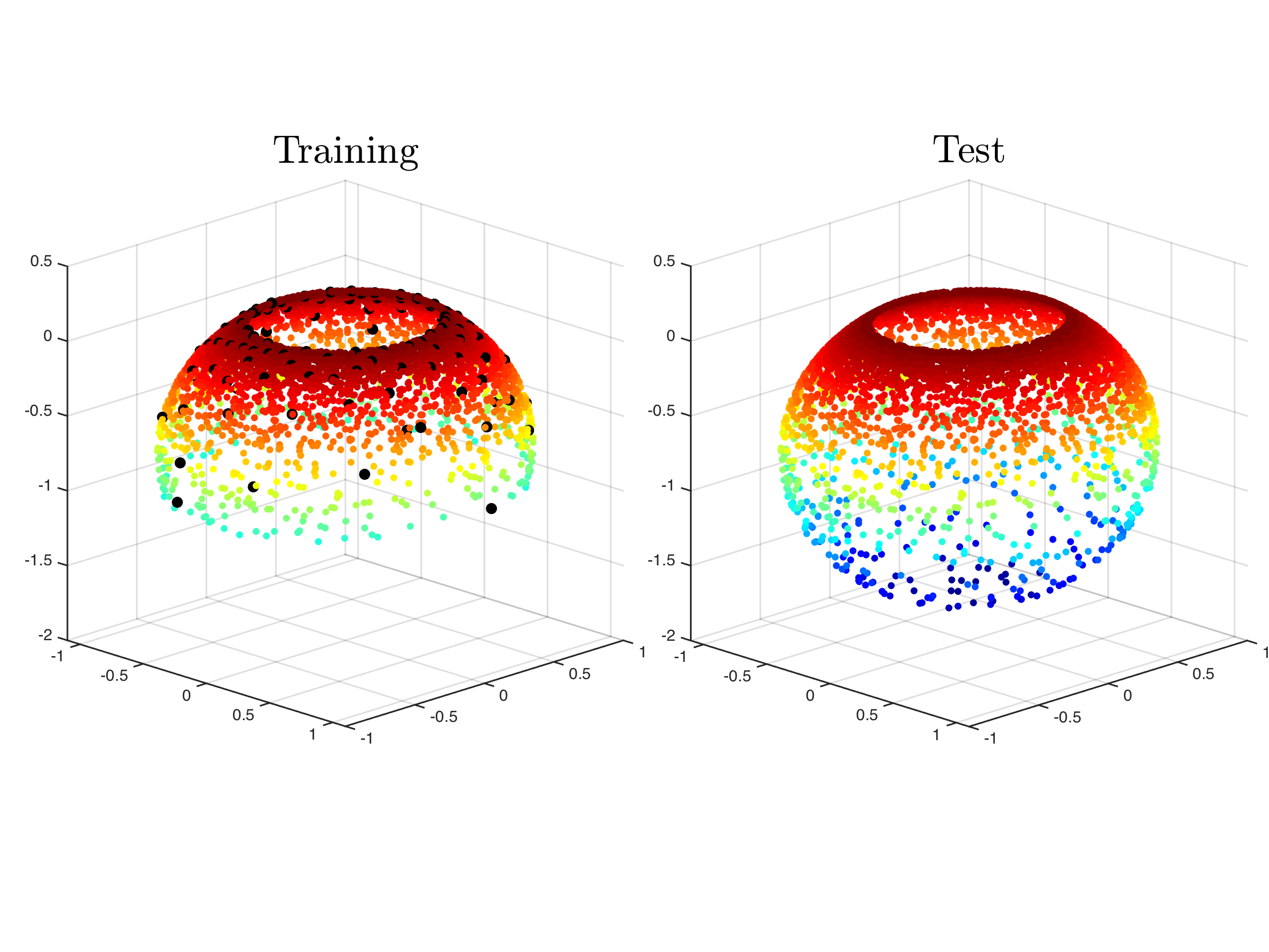}
\end{subfigure}
\vspace{-1cm}

\begin{subfigure}[h]{0.5\textwidth}
\includegraphics[scale=0.3]{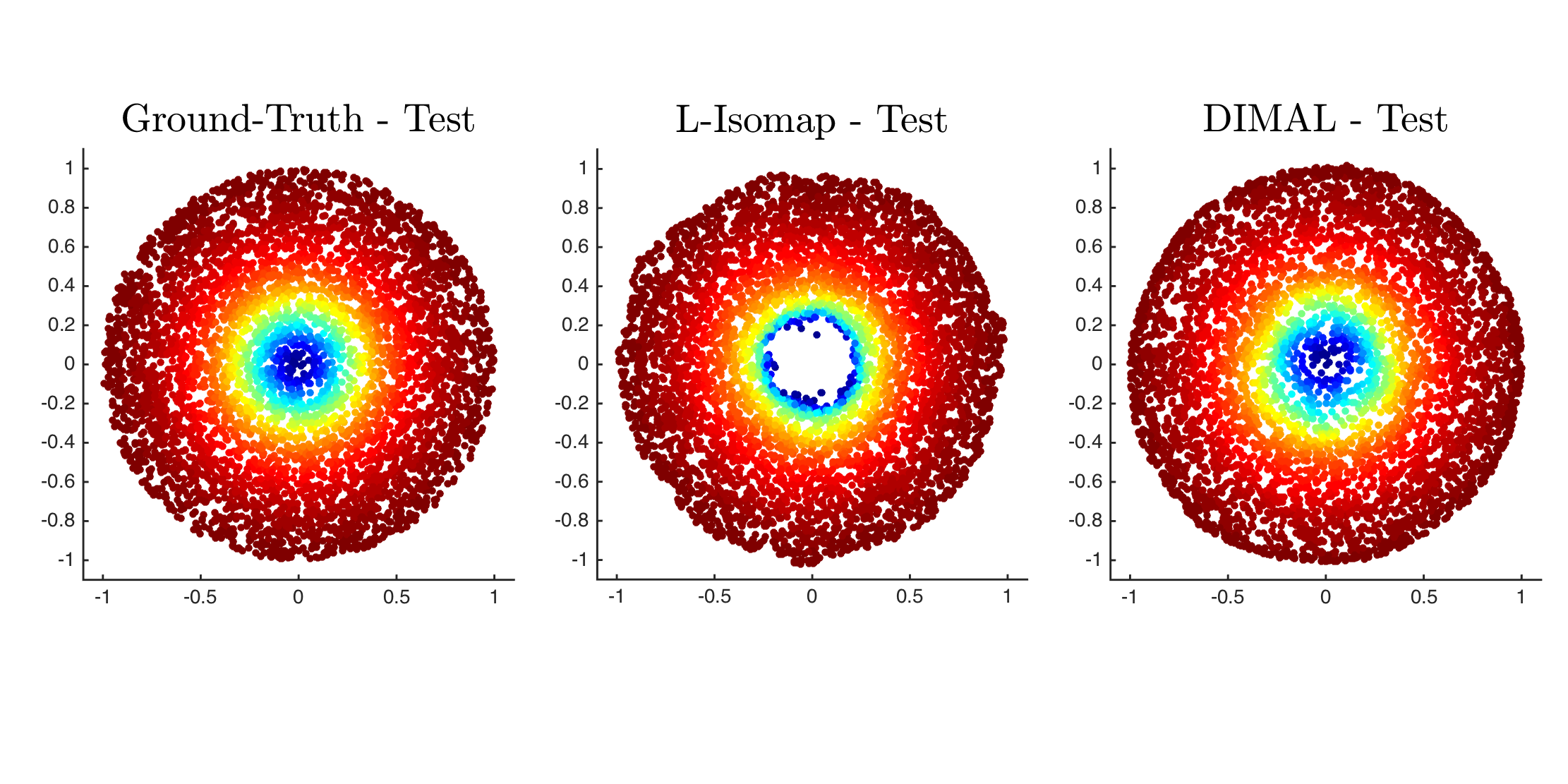}
\end{subfigure}
\end{center}
\vspace{-1cm}
{\caption{(a) The Conformal Fishbowl dataset \cite{silva2003global}, shown with the training and test cases. The training is done on landmarks obtained from only 60\% of the height of the fishbowl.
(b) The planar embeddings of Landmark Isomap and DIMAL.}
\label{fishbowl}}
\end{figure}

In \cite{silva2003global}, a technique called conformal isomap (C-Isomap) was proposed which modifies the local metric structure to account for an approximate conformal factor. C-Isomap is essentially an application of the classical scaling algorithm applied to a modified distance matrix which takes the conformal factor into account. A typical test case for this technique is the 3D conformal fishbowl visualized in Figure \ref{fishbowl}. The conformal fishbowl has a varying density along the manifold that is sparse at the bottom and dense near the rim of the bowl.  

We extend our DIMAL framework to this scenario using the same modified distances as proposed. To test the generalization abilities of both algorithms, we trained  on landmarks comprising from only 60\% of the height of the fishbowl and test on the flattening of the entire sample set. Figure \ref{fishbowl} demonstrates that DIMAL clearly provides a much better extension than Landmark Isomap which is unable to generalize to non-local points.

\subsubsection{Camera Pose Manifold}
\begin{figure}[t]
\begin{subfigure}[h]{0.23\textwidth}
\includegraphics[scale=0.1]{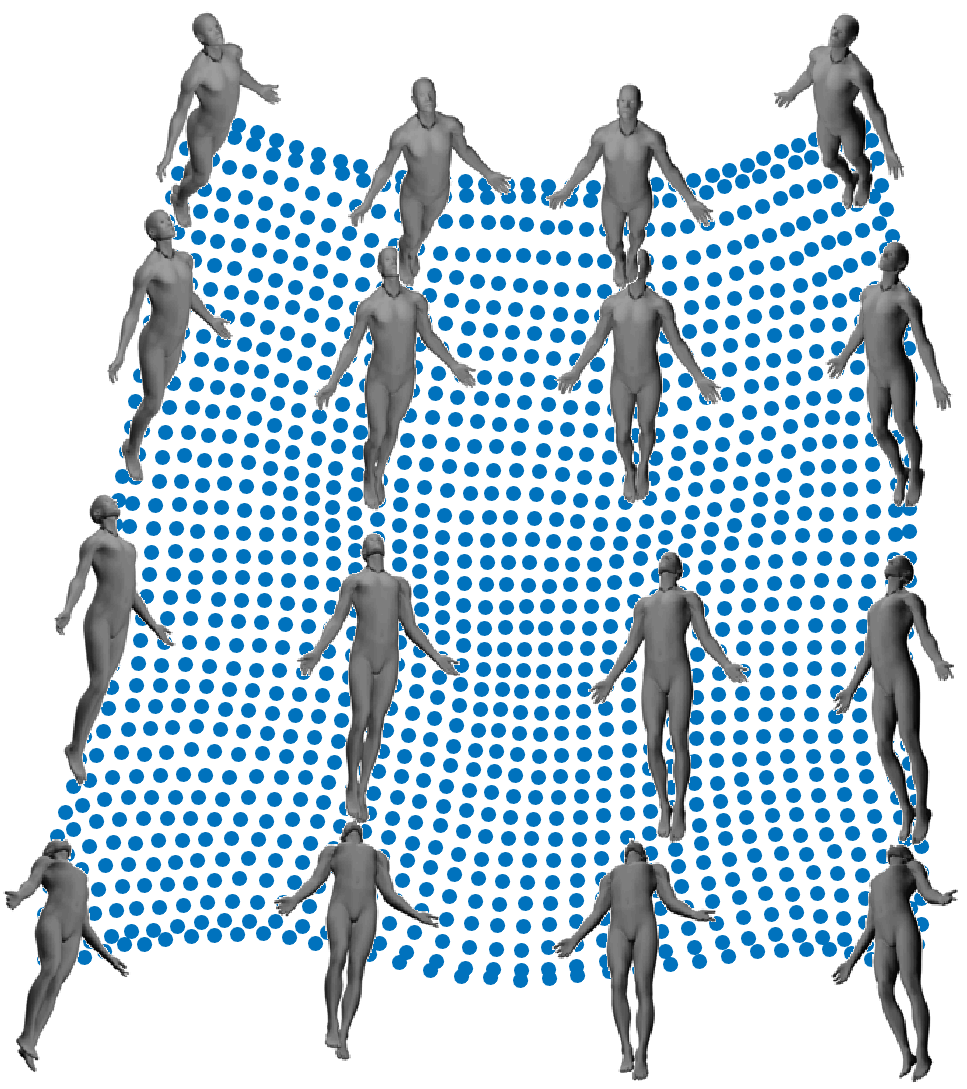}
\caption{DrLim.\\Training time $\approx$ 6620s}
\label{result_pose_DrLim}
\end{subfigure}
\begin{subfigure}[h]{0.23\textwidth}
\includegraphics[scale=0.1]{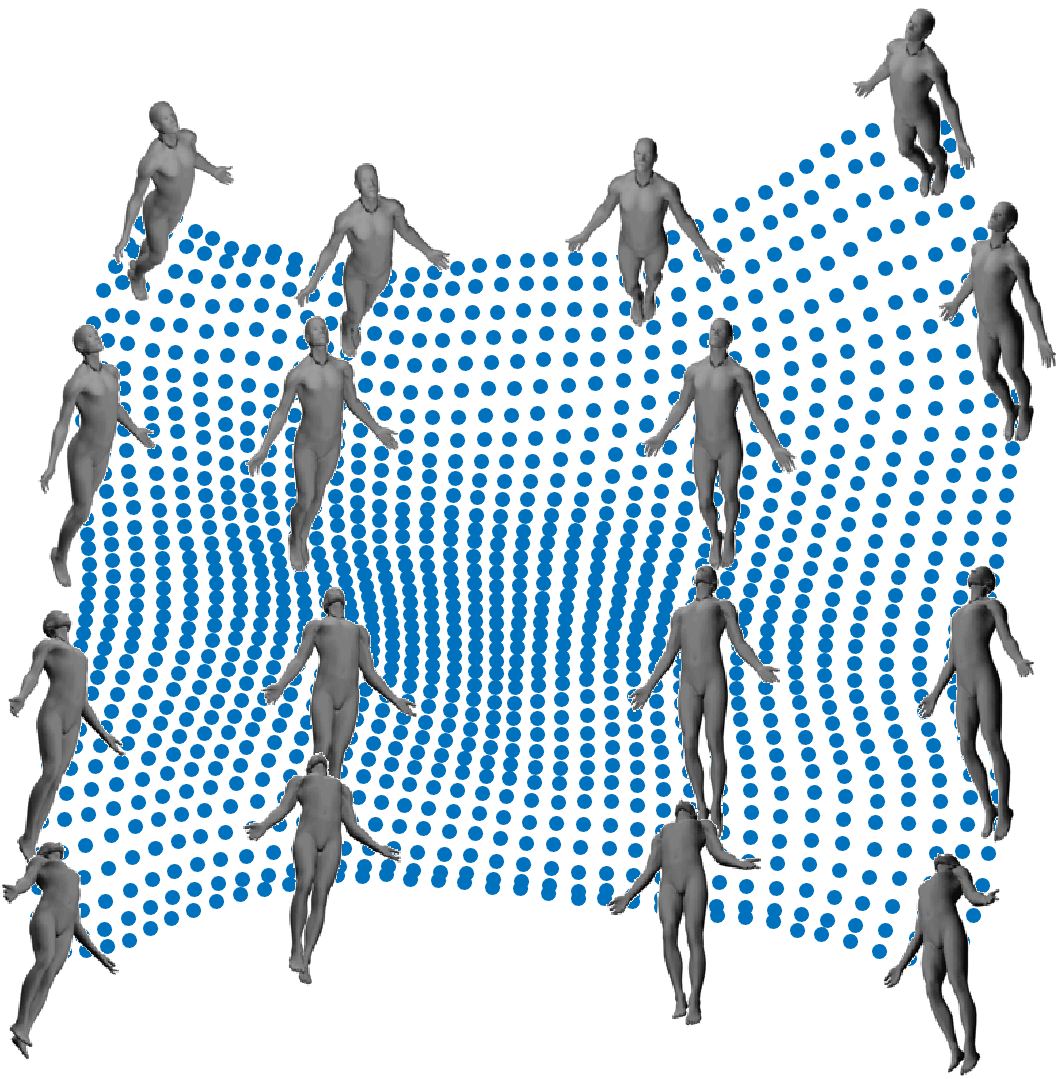}
\caption{DIMAL.\\Training time $\approx$ 860s}
\label{result_pose_MDSNet}
\end{subfigure}
{\caption{ {\bf Camera Pose Manifold:} Embedding results and training times for DrLim \cite{hadsell2006dimensionality} and DIMAL. DIMAL shows a faithful result for a much smaller training time.}
\label{result_pose}}
\end{figure}

Finally, we test our method on a more realistic dataset where the constraint of being isometric to a low-dimensional Euclidean space is not necessarily strict. We generate 1369 images obtained by smoothly varying the azimuth and elevation of the camera that is imaging a 3D object. We show in comparison, the visual results and associated training times of DrLim \cite{hadsell2006dimensionality} which was trained using the hinge-loss for the Siamese architecture of Figure \ref{siameseconfig}:  
\begin{equation}
\begin{aligned}
\mathcal{L}&(\Theta) =  \sum_{p} \;  \lambda^{(p)}\;||\mathcal{F}_{\Theta}(\mathbf{X}_{1}^{(p)}) - \mathcal{F}_{\Theta}(\mathbf{X}_{2}^{(p)})||\\
& + (1 -\lambda^{(p)})\;\max\;\{0,\; \mu - ||\mathcal{F}_{\Theta}(\mathbf{X}_{1}^{(p)}) - \mathcal{F}_{\Theta}(\mathbf{X}_{2}^{(p)})||\} 
\end{aligned}
\label{hinge_loss}
\end{equation}
DrLim requires a considerable training effort, requiring all possible $\binom{1369}{2} = 936396$ pairs whereas DIMAL yields a comparable result in a considerably smaller training time (geodesic distances between only $\binom{600}{2} = 179700$ pairs). We used the same architecture for DIMAL and DrLim for generating the embedding in Figures \ref{result_pose_DrLim} and \ref{result_pose_MDSNet}. DIMAL generates a comparable result with an order of magnitude smaller training time.
\section{Conclusion}
We explored an unsupervised deep learning approach to the isometric embedding problem. By training with a few landmarks we see that neural networks integrated into a classical framework like multidimensional scaling demonstrate improved generalization properties. The broad message of our paper is that, the somewhat \emph{black-box} learning infrastructures can be used for solving classical problems and this helps in developing new tools for better understanding their action. 

{\small
\bibliographystyle{ieee}
\bibliography{egbib}
}
\end{document}